\documentclass[letterpaper, 10 pt, conference]{ieeeconf}  %

\IEEEoverridecommandlockouts                              %

\usepackage{cite}
\usepackage{graphics} %
\usepackage{epsfig} %
\usepackage{mathptmx} %
\usepackage{times} %
\usepackage{amsmath} %
\usepackage{amssymb}  %
\usepackage{multirow}
\usepackage{booktabs} %
\usepackage{dsfont}
\usepackage{dcolumn}
\usepackage{tabularx}
\makeatletter
\let\NAT@parse\undefined
\makeatother
\usepackage[draft]{hyperref}
\usepackage{xcolor}
\makeatletter
\newcommand{\includegroupedpdf}[2]{{%
  \let\Gread@transgrouptrue\Gread@transgroupfalse
  \includegraphics[#1]{#2}}}
\makeatother
\newcommand{\method}{PrepTurn}

\hypersetup{hidelinks,pdfauthor={Zunzhi You, Yunke Wang, Haolan Guo, Chang Xu}}
\usepackage[font=footnotesize]{caption}
\DeclareCaptionLabelSeparator{periodspace}{.~~}
\renewcommand{\thetable}{\arabic{table}}
\newcommand{\topfloatguard}{\vspace*{6pt}}

\title{\LARGE \bf
Prepare to Turn: Learning Anticipatory Maneuvering\\
for Fast Sequential Quadrupedal Navigation
}
\author{Zunzhi You, Yunke Wang, Haolan Guo, Chang Xu%
\thanks{All authors are with School of Computer Science, University of Sydney.}%
}

\begin{document}

\maketitle
\thispagestyle{empty}
\pagestyle{empty}

\begin{abstract}
Fast quadrupedal navigation requires preparing for the next turn while approaching the current goal. Policies trained to reach and remain at a single target can instead produce stop-and-go motion when applied to sequential goals, compromising its agility potential. We introduce PrepTurn, a reinforcement learning framework for anticipatory maneuvering in sequential local navigation. PrepTurn maps the current and next pose goals directly to joint actions, using the upcoming goal to inform how the robot should approaches the current one. Its persistent-credit reward encourages timely completion of the whole sequence by continuing to reward reached goals while moving on and preparing subsequent maneuvers, and an automatic curriculum that gradually increases goal difficulty is critical for learning stable turning behaviors. Together, these components enable the policy to coordinate approach speed and reorientation across goal transitions without prescribed trajectories or arrival times. Simulation and onboard Unitree Go2 experiments demonstrate fast, efficient sequential traversal with high completion and reduced stop-and-go motion. We further show that when the goal-reaching tolerances are more relaxed, the policy learns to exploit the freedom allowed by the task with emergent behaviors, such as turning to the next goal in advance and cutting corners within the permitted regions behaviors, to traverse goal sequences more efficiently.

\end{abstract}
\suppressfloats[t]

\section{Introduction}

\begin{figure}[t]
\centering
\vspace{0.5em}
\includegraphics[width=0.99\linewidth]{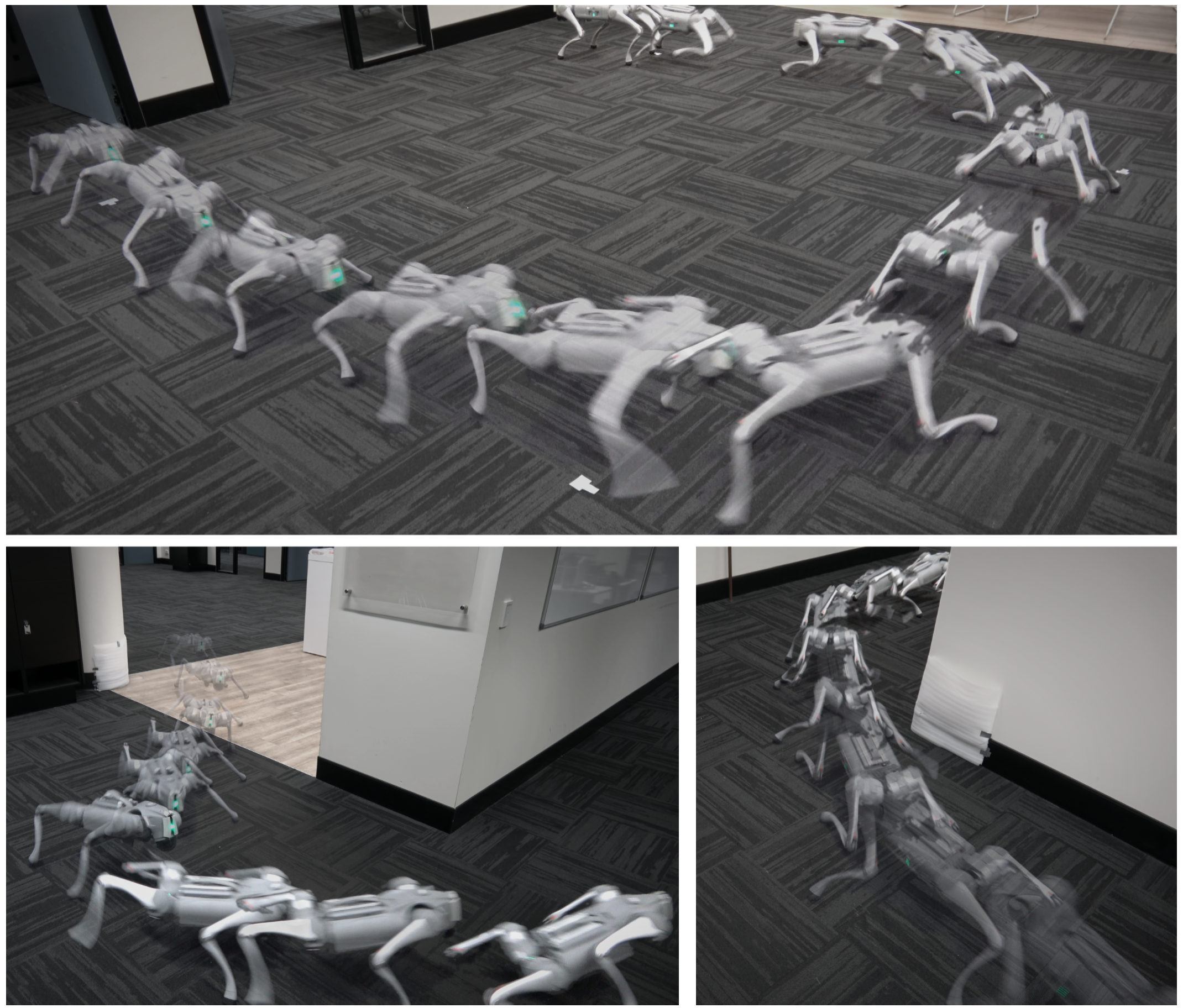}
\caption{Composited images of PrepTurn deployed on a Unitree Go2 in an indoor office. Given sequential pose goals, the policy coordinates its approach and turning motions through corridors and corners using onboard sensing and computation.}
\label{fig:teaser}
\end{figure}

Quadrupedal robots are especially useful as they can provide stable support in environments where wheeled platforms often struggle~\cite{tranzatto2022cerberus}. Many target applications of quadrupedal robots are time-sensitive. For example, in fire rescue~\cite{unitree_fire_rescue}, rapid reconnaissance and response can reduce risk to firefighters and save more lives and property. Even in routine service scenarios such as hospital delivery or campus logistics, efficient navigation through sequences of waypoints is highly desirable.

Fast sequential navigation requires coordinating how a robot arrives at a goal with where it must go next. 
Navigation in buildings and cluttered spaces involves frequent direction changes, which can be viewed as reaching a sequence of goals. 
For example, when leaving a room through a doorway, one needs to first reach the doorway (the initial goal), and then change the direction to navigate to a location outside the room (the subsequent goal).
The robot's approach speed, path, and heading adjustments should therefore account for the next goal before the current one is reached. 
Repeated stopping can increase traversal time, while carrying excessive speed into a sharp turn can lead to falls. 
We call this preparation \emph{anticipatory maneuvering}: future-goal information informs motion before the active goal switches. It can involve maintaining speed, slowing down, or reorienting within the current goal's constraints.

To date, most legged locomotion controllers are designed to track commanded base velocities~\cite{miki2022learning, zhuang2023robot, dialmpc2025}. 
This approach is effective for producing robust gaits, but it places the responsibility for when and how much to turn on an operator or an upstream planner that must generate velocity and yaw commands. 
More recently, goal-reaching approaches learn local navigation by conditioning the policy on a target goal, reducing the need for manual velocity commands~\cite{rudin2022advanced, zhang2024learning, he2024abs}. However, most existing formulations and evaluations focus on reaching a single terminal goal. They do not directly target the setting where the robot must traverse a \emph{sequence} of local goals at speed, where the primary challenge is the direction change between successive goals.
  
In this work, we formulate the \textit{sequential local navigation} task as traversing an ordered sequence of pose goals, and introduce \textbf{PrepTurn}, a reinforcement learning framework for this setting.
PrepTurn maps proprioception and the current and next pose goals directly to joint actions, so the upcoming maneuver can shape motion before the current goal is reached.
We further design a persistent-credit reward that retains the value of completed goals and encourages reliable traversal through the whole goal sequence. 
We also adopt an automatic curriculum that progressively expands the turning difficulty of sampled goal sequences.

We evaluate PrepTurn on randomized two- and five-goal sequences, comparing it with guided velocity tracking and single goal reaching. PrepTurn achieves faster traversal than both baselines while maintaining high success rates. We further shows how reaching tolerances shape turning behavior and traversal efficiency, and use ablations to assess the roles of reward design, lookahead, and curriculum. Onboard Unitree Go2 trials demonstrate sequential traversal with shorter mean traversal times than single goal reaching. 
To summarize, our contributions include:
\begin{itemize}
\item The formulation of the sequential local navigation task to evaluate the crucial ability of reaching ordered pose goals while preparing for subsequent maneuvers during high-speed locomotion.
\item PrepTurn, a reinforcement learning framework combining a novel sequential goal-reaching reward, next-goal lookahead, and a goal curriculum to learn anticipatory control across sequential goals.
\item Extensive validation in simulation comparisons, tolerance studies, component ablations, and onboard deployment on a Unitree Go2 that demonstrates PrepTurn's fast sequential local navigation and emergent behaviors.
\end{itemize}

\section{Related Work}
\label{sec:related}

\subsection{Agile Locomotion and Maneuvering}

Agile legged locomotion has been studied through model-based optimization~\cite{grandia2023perceptive,di2018dynamic,dialmpc2025} and reinforcement learning~\cite{ha2025survey,rudin2022learning}.
Learned controllers have demonstrated robust terrain and obstacle traversal~\cite{miki2022learning, zhuang2023robot, he2025attention, cheng2024extreme, kang2026agile} and have also enabled more agile skills, such as continuous jumping
~\cite{yang2023cajun, yang2025agile} and fall recovery~\cite{ma2023learning, wang2024guardians}.

Closely related to our setting are high-speed running and maneuvering while moving. High forward speeds have been achieved using both model predictive control (MPC)~\cite{park2017high, kim2019highly} and reinforcement learning
~\cite{jin2022high, bellegarda2022robust, margolis2024rapid},
but these methods primarily aim to maximize straight-line speed rather than to handle changes of direction. Other works study maneuverability, including spinning in place~\cite{mastalli2022agile, margolis2024rapid}, twisting jumps~\cite{zhou2022momentum}, and 
 turning while moving~\cite{yu2022dynamic, han2024lifelike}. 
Some of these methods exploit additional effectors, such as active spines~\cite{li2026scheetah} or using a manipulator as a tail to manage angular momentum during turning~\cite{huang2024manipulator, yang2025learning}. 
These capabilities motivate studying how running and turning should be coordinated during navigation.
In a velocity-command interface, the timing and magnitude of directional changes are specified upstream of the locomotion controller.
Our focus is instead a local policy that directly uses sequential pose goals to coordinate approach and reorientation, rather than relying on an upstream module to specify the velocity commands for each maneuver.

\subsection{Goal-Conditioned Quadrupedal Navigation}
\label{sec:goal-reaching}

Goal-conditioned policies extend locomotion beyond velocity tracking by directly pursuing spatial targets.
They have enabled obstacle traversal~\cite{rudin2022advanced}, challenging terrain negotiation~\cite{zhang2024learning,xu2024dexterous,zhang2025motion}, collision avoidance~\cite{he2024abs,zhong2025bridging}. These capabilities are realized either in an end-to-end fashion, or via hierarchical architectures where a planner provides local goals~\cite{hoeller2024anymal} or foothold plans~\cite{kim2025high} to a low-level locomotion controller. 
These goal-reaching formulations motivate extending local control to successive goals. 
RobotKeyframing~\cite{zargarbashi2024robotkeyframing} conditions behavior on multiple future pose or posture targets.
However, this approach requires assigning prescribed arrival times to each intermediate target and penalizes deviations from the exact keyframe timings.
Our task instead specifies ordered pose goals without per-goal arrival timestamps or mandatory intermediate stops, and evaluates traversal time under reaching constraints.
We therefore focus on how future-goal context and sequential goal-reaching incentives enable a local policy to prepare for the next maneuver while approaching the current goal, coordinating reorientation and momentum across goal transitions.

\begin{figure*}[t]
    \centering
    \topfloatguard
    \includegroupedpdf{width=0.92\linewidth}{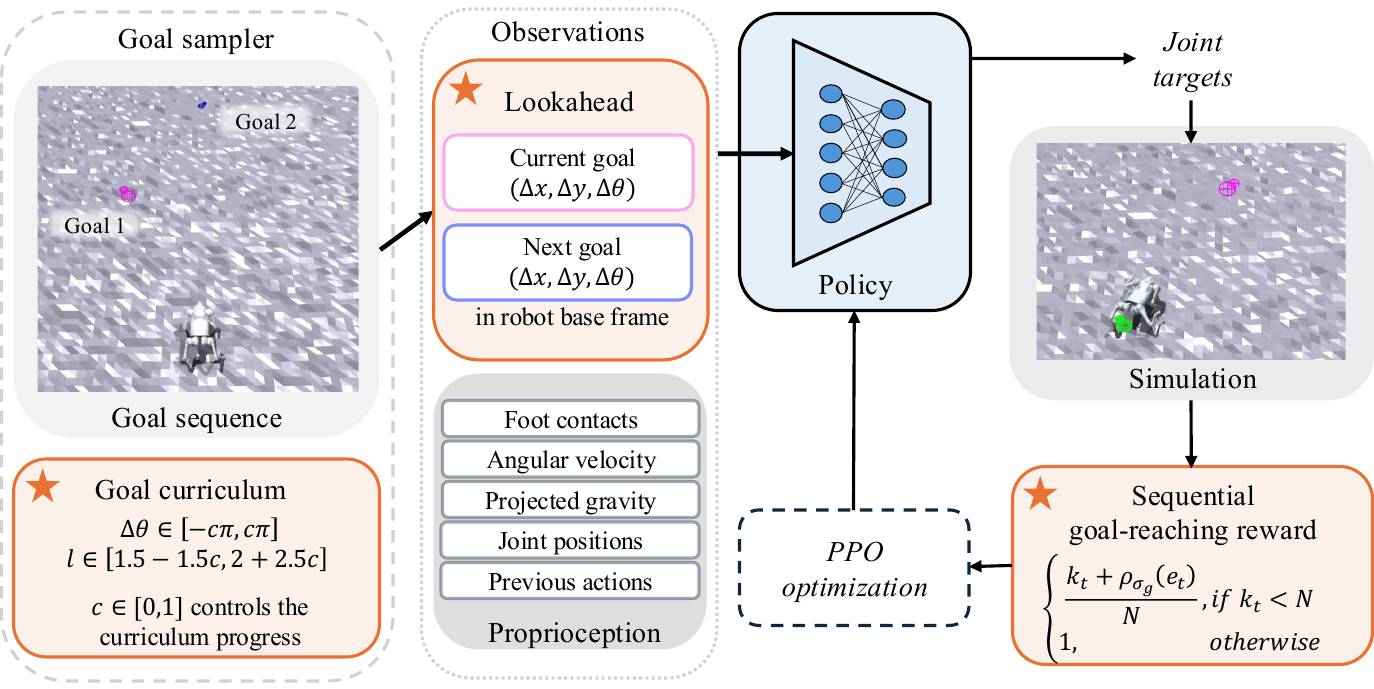}
    \caption{Overview of the \method{} framework. The Goal Sampler generates a sequence of goals based on the curriculum. The Command Updater advances the goal index upon goal reaching and provides the pose of current and future goals in the robot base frame. The Policy takes the commands and proprioceptive states as input to output joint position targets.}
\label{fig:method}
\end{figure*}

\section{Sequential Local Navigation}
\label{sec:formulation}

In this section, we first review the single-goal local navigation formulation used in prior work~\cite{rudin2022advanced, he2024abs}, then extend it to the \textit{sequential local navigation} task where a robot must traverse an ordered sequence of goals.

\subsection{Background: Single-Goal Local Navigation}
\label{sec:background}

Recent work on agile legged locomotion formulates the control problem as goal-conditioned local navigation~\cite{rudin2022advanced, zhang2024learning, he2024abs}.
Instead of tracking a commanded base velocity, the robot is given a target pose and must reach it within a fixed-length episode.
We briefly review the formulation of~\cite{rudin2022advanced}, which provides the single-goal formulation that we extend.

\noindent \textbf{Observation and action spaces.}
The policy receives an observation $o_t$ consisting of proprioceptive states, goal commands, and the remaining episode time $T-t$.
The proprioceptive states include foot contact indicators $c_f\in\{0,1\}^{4}$, base angular velocity $\boldsymbol{\omega}\in\mathbb{R}^{3}$, projected gravity in the base frame $\mathbf{g}\in\mathbb{R}^{3}$, joint positions $\mathbf{q}\in\mathbb{R}^{12}$, joint velocities $\dot{\mathbf{q}}\in\mathbb{R}^{12}$, and the previous action $\mathbf{a}_{t-1}\in\mathbb{R}^{12}$.
The goal command $G^c\in\mathbb{R}^{3}$ encodes the relative position $(\Delta x, \Delta y)$ and heading $\Delta\theta$ of the goal in the robot's base frame.
The action $\mathbf{a}_t\in\mathbb{R}^{12}$ specifies the target joint positions.

\noindent \textbf{Single-goal-reaching reward.}
The single-goal-reaching reward used in~\cite{rudin2022advanced, he2024abs} encourages the robot to reach and remain at a single target near the end of the episode. Its tracking terms share the following form:
\begin{equation}
\label{eq:single_goal_reward}
r_{\text{track}} = \frac{1}{1 + \left\| \frac{e}{\sigma} \right\|^2} \cdot \frac{\mathds{1}(t > T - T_r)}{T_r},
\end{equation}
where $e$ is the tracking error (e.g., the Euclidean distance to the goal position, or the heading error), $\sigma>0$ is a constant controlling the kernel's sharpness, $T$ is the total episode length, and $T_r$ is a time threshold.
The indicator function $\mathds{1}(t > T - T_r)$ restricts the reward strictly to the final $T_r$ seconds of the episode, normalized by $1/T_r$.

This design frees the robot from explicit velocity constraints during earlier phases of the episode: it simply needs to arrive at the goal before $T - T_r$ and remain there to maximize the reward~\cite{rudin2022advanced}.
Additional task rewards (e.g., an agility bonus and a stall penalty) and regularizations (penalizing large torques, joint accelerations, etc.) complete the reward function.
However, because the reward heavily prioritizes stable standing at the terminal goal, the policy is inherently incentivized to decelerate upon approach.
Therefore, when this policy is deployed for sequential local navigation by updating its target after each goal, its learned deceleration can persist. Moreover, an observation containing only the current goal does not reveal the upcoming turn. These objective and information differences motivate training directly for sequential traversal.

\subsection{Task Formulation}
\label{sec:task_formulation}

We extend the single-goal setting to \textit{sequential local navigation}, where the robot must traverse an ordered goal sequence $\mathcal{G}=\{g_1,\dots,g_N\}$.

\noindent \textbf{Goal indexing and switching.}
Let $k_t \in \{0,\dots,N\}$ denote the number of goals already reached by time step $t$.
If $k_t < N$, the current goal is $g_{k_t+1}$. Once the reaching condition for $g_{k_t+1}$ is satisfied, the counter increments by one (up to $N$).

For goals defined by $\mathrm{SE}(2)$ poses, $g=(x_g,y_g,\theta_g)$, the reaching condition is bounded by two thresholds: $\epsilon_{xy}$ for position and $\epsilon_{\theta}$ for heading. Given the robot's planar pose $(x_t,y_t,\theta_t)$ at time step $t$, the goal is reached if:
\begin{equation}
\label{eq:pose_reach}
\bigl\|(x_t,y_t)-(x_g,y_g)\bigr\|_2 < \epsilon_{xy}
\quad \land \quad
\bigl|\mathrm{wrap}(\theta_t-\theta_g)\bigr| < \epsilon_{\theta},
\end{equation}
where $\mathrm{wrap}(\cdot)$ maps angles to $(-\pi,\pi]$.

In practice, we find that relying solely on this condition is brittle for existing single-goal-reaching methods. Without meticulous tuning, the thresholds are either too restrictive or too loose, causing the robot to stop near a goal without ever formally reaching it, or encouraging the robot to turn at high speeds.
To mitigate this, we introduce a supplementary reaching condition. This condition employs relaxed positional and heading thresholds ($\epsilon_{xy}^+$ and $\epsilon_{\theta}^+$) but strictly triggers only when the robot is nearly stationary, defined by limits on its base planar linear velocity $v$ and angular velocity $\omega$:
\begin{equation}
\label{eq:pose_reach_static}
\begin{aligned}
\bigl\|(x_t,y_t)-(x_g,y_g)\bigr\|_2 &< \epsilon_{xy}^+ \quad\land\quad \bigl|\mathrm{wrap}(\theta_t-\theta_g)\bigr| < \epsilon_{\theta}^+ \\
\land\quad v &< 0.1~\text{m/s}
\quad\land\quad \omega < 0.1~\text{rad/s}.
\end{aligned}
\end{equation}
Unless otherwise specified, our experiments use $(\epsilon_{xy},\epsilon_{\theta})=(0.1\,\mathrm{m},5^\circ)$ and $(\epsilon_{xy}^+,\epsilon_{\theta}^+)=(0.5\,\mathrm{m},60^\circ)$.

\section{PrepTurn: Learning Anticipatory Maneuvering}

\subsection{Sequential Goal-Reaching Reward}
\label{reward}

The terminal objective in Section~\ref{sec:background} rewards reaching and remaining at one goal. For sequential local navigation, a natural starting point is to reward progress toward the active goal and give a one-time credit when it is reached. We first define the active-goal pose error used by both reward formulations. Let $d_{xy} = \bigl\|(x_t,y_t)-(x_g,y_g)\bigr\|_2$ and $d_{\theta} = \bigl|\mathrm{wrap}(\theta_t-\theta_g)\bigr|$. To down-weight heading error far from the goal, we define
\begingroup
\begin{align}
    \label{eq:pose_error}
    e_t = d_{xy} + \lambda_{\theta}\,\frac{d_{\theta}}{1+(d_{xy}/\sigma_{\theta})^2}.
\end{align}
\endgroup

Here $\lambda_{\theta}$ weights the heading term, and $\sigma_{\theta}>0$ sets the distance scale over which that term becomes influential. Unlike prior work that rewards position and heading tracking separately~\cite{rudin2022advanced, he2024abs}, our pose error couples heading alignment to position proximity. This makes heading error matter mainly near the goal, avoiding pressure to align with the goal heading too early and then approach sideways or backward.

The \emph{simple-progress reward} provides signed reduction in this error and a one-time credit for each reached goal. Let $k_t^-$ and $k_t^+$ be the completed-goal counts before and after a control step, and let $e_t^-$ and $e_t^+$ be the errors at these two states relative to the \emph{same pre-step active goal}. With $\Delta k_t=k_t^+-k_t^-$, its reward rate is
\begingroup
\begin{equation}
\label{eq:progress_reward}
r^{\mathrm{prog}}_t =
\begin{cases}
\dfrac{(e_t^- - e_t^+)/\ell_0+\Delta k_t}{N\Delta t}, & k_t^-<N,\\[6pt]
1, & k_t^-=N,
\end{cases}
\end{equation}
\endgroup
where $\ell_0=3$\,m and $\Delta t=0.02$\,s. Negative progress is retained, and error differences never span a goal switch or episode reset. This reward pays for moving closer to the active goal and gives each completion a single bonus before moving on to the next goal.

Simple progress supports fast traversal but yields more falls in our experiments (Section~\ref{sec:reward_factorial}). We therefore design a \emph{persistent-credit reward} that combines a proximity kernel with continuing credit for completed goals:
\begingroup
\begin{align}
\label{eq:kernel}
\rho_{\sigma_g}(e) \;=\; \frac{1}{1+\bigl(\tfrac{e}{\sigma_g}\bigr)^2},
\end{align}
\begin{align}
\label{eq:sequential_reward}
r^{\text{seq}}_t \;=\;
\begin{cases}
\dfrac{k_t^+ + \rho_{\sigma_g}(e_t)}{N}, & k_t^+<N,\\[6pt]
1, & k_t^+=N.
\end{cases}
\end{align}
\endgroup
Here $\sigma_g>0$ sets the kernel's pose-error scale, and $e_t$ is evaluated after the step against the active goal after any goal switch.

The two terms address complementary needs. Near the current goal, preparing for the next maneuver may require slowing down, reducing the rate of current-goal error reduction. When the pose error is small relative to $\sigma_g$, the kernel becomes less sensitive to further error reduction, easing the pressure to maintain approach speed during turn preparation.

The completed-goal term shifts emphasis toward goal completion: each reached goal adds $1/N$ to the reward rate on every subsequent step, regardless of current-goal error. Completing each goal earlier yields more accumulated credit over the remaining episode, encouraging timely completion of successive goals.

\subsection{Goal Generation and Curriculum}
\label{sec:goal_curriculum}

\noindent \textbf{Goal generation.}
During training, the Goal Sampler (Fig.~\ref{fig:method}) generates an episode goal sequence $\mathcal{G}=\{g_1,\dots,g_N\}$.
The first goal $g_1$ is sampled relative to the robot start pose; each subsequent goal $g_{i>1}$ is sampled relative to the previous goal pose.
In the local reference frame, we first advance a 0.5\,m pre-step along the reference heading to ensure forward motion and avoid degenerate near-goal cases. We then sample a signed turning offset $\Delta\theta$ and a travel length $\ell$ from the active curriculum ranges, and set
\begin{equation}
\theta_g = \mathrm{wrap}(\theta_{\mathrm{ref}} + \Delta\theta), \qquad
\mathbf{p}_g = \mathbf{p}_{\mathrm{ref}} + 0.5\,\mathbf{h}(\theta_{\mathrm{ref}}) + \ell\,\mathbf{h}(\theta_g),
\end{equation}
where $(x_{\mathrm{ref}},y_{\mathrm{ref}},\theta_{\mathrm{ref}})$ denotes the reference planar pose,
$\mathbf{p}_{\mathrm{ref}}=(x_{\mathrm{ref}},y_{\mathrm{ref}})$ and $\mathbf{p}_g=(x_g,y_g)$, and $\mathbf{h}(\theta)=(\cos\theta,\sin\theta)$.

\noindent \textbf{Goal curriculum.}
We implement an automatic curriculum that rescales the sampling ranges of $(\Delta\theta,\ell)$ based on the goal-reaching success rate over a rolling window.
Each reached goal counts as a success, whereas any remaining unreached goals at sequence termination are recorded as failures.
Periodically, we update a scalar curriculum progress variable $c\in[0,1]$: if the current success rate exceeds an expansion threshold (80\%), we increase $c$ to heighten difficulty; if it drops below a contraction threshold (20\%), we decrease $c$.
We then linearly interpolate the sampling ranges between an easy regime (straight, short goals: $\Delta\theta=0$, $\ell\in[1.5,2.0]$\,m) and a hard regime
(large turns with variable distances: $\Delta\theta\in[-\pi,\pi]$, $\ell\in[0,4.5]$\,m).

\subsection{Lookahead for Maneuvering Preparation}

Preparation requires knowing the upcoming maneuver before the active goal switches. We therefore provide a short lookahead window containing the current and upcoming goals, so that the policy can condition the current approach on the next target.
Based on the goal counter $k_t$, the $i$-th goal in the lookahead window is defined as:
\begin{equation}
\label{eq:lookahead_index}
\hat{g}_{t,i} = g_{\min(k_t+i+1,\,N)}, \qquad i\in\{0,\dots,n-1\},
\end{equation}
where $n$ is the number of goals included in the observation.
To maintain a constant observation dimension when fewer than $n$ goals remain, we duplicate the final goal (e.g., after reaching $g_{N-1}$, the lookahead window repeats $g_{N}$).

The next goal reveals the upcoming direction change while the current goal remains active. The policy can use this information to adjust its approach speed and, where the current goal's reaching tolerances permit, its heading and path.

\section{Experimental Results}
In this section, we demonstrate the effectiveness of the proposed method by comparing the performance on sequential local navigation tasks with two baseline methods, and examine how goal-reaching tolerances shape traversal behavior. We further ablate the reward, lookahead, and curriculum, and evaluate onboard deployment on a Unitree Go2.

\subsection{Experimental Setup}
\label{sec:setup}

All policies are trained with PPO~\cite{schulman2017proximal} in the Isaac Gym~\cite{makoviychuk2021isaac} GPU-based simulator using the Unitree Go2 robot model. We evaluate sequential local navigation on randomized rough terrain, where the robot must reach an ordered sequence of $N$ pose goals, each specifying a target position and heading.
We consider two tasks: a two-goal task ($N=2$) involving a single turn and a five-goal task ($N=5$) involving successive turns. Both tasks sample turn angles in $[-\pi,\pi]$. Goal spacing is fixed at 3\,m for $N=2$ and sampled in $[1.5,3.0]$\,m for $N=5$, with evaluation horizons of 12\,s and 24\,s, respectively. PrepTurn uses the persistent-credit reward and observes the current and next pose goals.
For each method, policies trained with three different random seeds are evaluated on the same $M=4096$ goal sequences per task and a shared rough-terrain realization.

\noindent\textbf{Metrics.}
Success rate (SR) is the percentage of episodes that reach all goals within the evaluation horizon without a terminal fall. Fall rate (FR) is the percentage of episodes terminated by a fall. Both rates use all evaluation episodes as the denominator. All remaining episodes are timeouts, accounting for $100\%-\mathrm{SR}-\mathrm{FR}$. To include failures when comparing speed, we report
\begingroup
\begin{equation}
\bar t_{\rm FA}=\frac{1}{M}\sum_{i=1}^{M}
\left[c_i t_i+(1-c_i)H\right],
\end{equation}
\endgroup
where $c_i$ indicates completion of episode $i$, $t_i$ is its completion time, and $H$ is the evaluation horizon; failures receive the full horizon rather than appearing artificially fast. Failure-adjusted time is comparable only within a fixed task and horizon. $E_{xy}$ is the conditional closest approach to each goal while it is active; never-activated future goals are excluded, so it must be read together with the outcome partition rather than as unconditional route accuracy. To characterize transition behavior rather than task outcome, we additionally report planar speed $v_g$ at goal switches.

\subsection{Comparison with Baselines}
\label{sec:baseline}

We compare our PrepTurn with two baselines. First, we implement \emph{guided velocity tracking}, a modular baseline in which a geometric command generator converts the same current-and-next-goal input as PrepTurn into velocity commands for a velocity-tracking locomotion policy trained with PPO. Planar velocity points toward the active goal, while the yaw-rate command turns the robot toward the active goal's required heading. At intermediate goals, the commanded speed is adjusted according to the angle from the active goal's required heading to the next segment $\Delta\theta$: $s=v_{\max}(1-\tfrac12 b(d)|\Delta\theta|/\pi)$, where $v_{\max}$ is set to $1.5\,m/s$, and the weight $b(d)$ rises smoothly from zero to one as the goal distance $d$ decreases from 0.5\,m to zero, and is zero farther away. This slows sharper turns without prescribing intermediate stops. We also conducted experiments with $v_{\max}$ set to $1.0\,m/s$ and $2.0\,m/s$, and removing anticipatory slowdown, resulting in either substantially increased failure-adjusted time or increased FR.

The other baseline is a goal-conditioned joint-action policy adapted from the terminal objective in~\cite{he2024abs}, termed as \emph{single goal reaching}. It observes only the active pose goal and is trained to reach and remain at that goal using the tracking reward in Eq.~\eqref{eq:single_goal_reward}, which rewards small position and heading errors during the final $T_r$ seconds of the episode. For sequential evaluation, we replace its target whenever a goal is reached. It uses the same goal generator and curriculum as PrepTurn.

\begin{table*}[t]
\centering
\caption{Sequential local navigation with randomized goal sequences. Values are mean $\pm$ sample SD across three training seeds.}
\label{tab:randomized}
\small\setlength{\tabcolsep}{3pt}
\renewcommand{\arraystretch}{1.12}
\begin{tabular*}{\textwidth}{@{\extracolsep{\fill}}l cc ccc@{}}
\toprule
& \multicolumn{2}{c}{Outcomes (\%)} & & & \\
\cmidrule(lr){2-3}
Method & $\mathrm{SR}\uparrow$ & $\mathrm{FR}\downarrow$ & $t_{\rm FA}\downarrow$ (s) & $E_{xy}\downarrow$ (m) & $v_g$ (m/s)\\
\midrule
\multicolumn{6}{@{}l}{\emph{$N=2$, $H=12$\,s}}\\
Guided velocity tracking & $96.06\pm2.36$ & $3.17\pm1.79$ & $6.02\pm0.10$ & $0.15\pm0.02$ & $0.71\pm0.05$\\
Single goal reaching & $99.89\pm0.11$ & $0.11\pm0.11$ & $4.40\pm0.17$ & $0.06\pm0.03$ & $0.09\pm0.02$\\
\textbf{PrepTurn} & $99.87\pm0.16$ & $0.11\pm0.18$ & $3.20\pm0.14$ & $0.09\pm0.00$ & $1.42\pm0.08$\\
\midrule
PrepTurn (w/o lookahead) & $99.89\pm0.04$ & $0.02\pm0.02$ & $4.71\pm0.17$ & $0.09\pm0.01$ & $0.89\pm0.13$\\
PrepTurn (simple progress) & $97.69\pm1.67$ & $2.29\pm1.70$ & $3.17\pm0.10$ & $0.11\pm0.02$ & $2.31\pm0.13$\\
\midrule\midrule
\multicolumn{6}{@{}l}{\emph{$N=5$, $H=24$\,s}}\\
Guided velocity tracking & $82.00\pm5.31$ & $14.83\pm5.06$ & $13.30\pm0.53$ & $0.14\pm0.01$ & $0.99\pm0.03$\\
Single goal reaching & $99.37\pm0.26$ & $0.63\pm0.26$ & $9.22\pm0.25$ & $0.06\pm0.01$ & $0.09\pm0.02$\\
\textbf{PrepTurn} & $99.43\pm0.34$ & $0.49\pm0.20$ & $6.65\pm0.26$ & $0.09\pm0.00$ & $1.33\pm0.06$\\
\midrule
PrepTurn (w/o lookahead) & $98.49\pm0.88$ & $0.76\pm0.87$ & $10.07\pm0.07$ & $0.10\pm0.01$ & $0.72\pm0.12$\\
PrepTurn (simple progress) & $96.50\pm2.45$ & $3.47\pm2.41$ & $6.69\pm0.37$ & $0.09\pm0.01$ & $2.10\pm0.18$\\
\bottomrule
\end{tabular*}
\end{table*}

Table~\ref{tab:randomized} shows that guided velocity tracking does not reliably complete every goal in a sequence, particularly for $N=5$, where its success rate is only $82.00\%$. Although its command generator adjusts speed according to the upcoming change in direction, this lookahead-based rule is insufficient for reliable traversal, suggesting that more refined command design or a learned command generator is needed. Its failure-adjusted time is also substantially longer than that of both goal-reaching methods: $6.02$\,s at $N=2$ and $13.30$\,s at $N=5$, compared with $4.40$/$9.22$\,s for single goal reaching and $3.20$/$6.65$\,s for PrepTurn. Thus, the tested guidance rule does not realize the velocity tracker's potential for agile sequential traversal. Its higher conditional position error ($E_{xy}=0.14$--$0.15$\,m, versus $0.06$\,m and $0.09$\,m, respectively) also indicates less precise approaches to active goals.

Single goal reaching completes the sequences accurately and reliably, with success rates above $99\%$ and the lowest $E_{xy}$ in both tasks. However, its failure-adjusted time is approximately $38\%$ longer than PrepTurn's. This difference is consistent with its objective of reaching and remaining at each target: the robot approaches a near stop before the target switches, then accelerates toward the next goal. Its low goal-switch speed, $v_g=0.09$\,m/s in both tasks, supports this stop-and-go interpretation. Fig.~\ref{fig:tolerances} compares trajectories on four fixed courses, with color indicating base speed. In its first column, single goal reaching slows to near-zero speed around the goals and regains speed between them, qualitatively illustrating the stop-and-go behavior associated with its longer traversal times in Table~\ref{tab:randomized}. PrepTurn instead maintains $v_g=1.42$\,m/s at $N=2$ and $1.33$\,m/s at $N=5$, carrying motion through goal transitions while retaining success rates above $99\%$. The comparison illustrates how learning to traverse a sequence can improve traversal efficiency over repeatedly solving a terminal reaching objective, with a modest increase in conditional position error.

\subsection{Effect of Goal-Reaching Tolerances}
\label{sec:tolerances}

In the formulation of the sequential local navigation task, the thresholds of the reaching conditions determines when a goal is considered to be reached, which reflect the requirements of different applications and environments. For example, when passing through a doorway, we generally do not need to prescribe the heading as long as the robot is close enough to the goal and able to get itself through. Moreover, the doorway could be wide enough that the robot can be 0.5\,m away from the goal and still be able to pass through.

Such tolerances define the freedom available for preparing the next maneuver, and we expect relaxed requirements to enable more efficient traversal.
Without a heading requirement, the robot can begin turning toward the next goal before reaching the current goal, rather than first aligning with a prescribed heading. A wider position tolerance additionally allows it to cut corners within the accepted goal region, taking a shorter route toward the next goal.

\begin{figure}[t]
\centering
\includegraphics[width=\columnwidth]{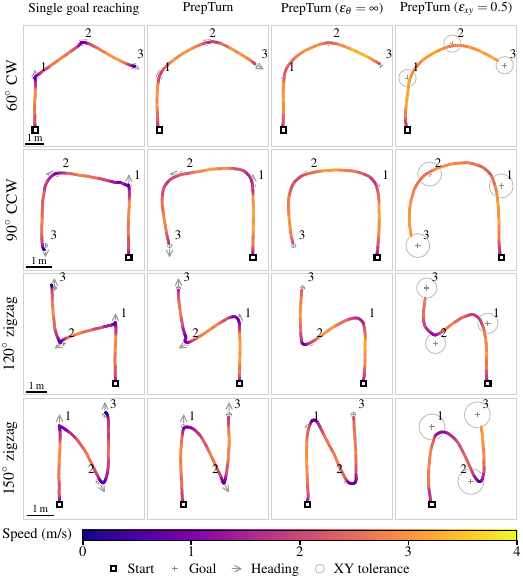}
\caption{Trajectories across methods and courses. Color indicates base speed; circles and arrows show position tolerances and required headings.}
\label{fig:tolerances}
\end{figure}

Fig.~\ref{fig:tolerances} illustrates these learned behaviors on four fixed courses, comparing single goal reaching with three PrepTurn policies: position-and-heading reaching, position-only reaching, and wider position-only reaching. Direct-switch tolerances are 0.1\,m in position and $5^\circ$ in heading for the first two columns. The last two columns remove the heading requirement ($\epsilon_\theta=\infty$); the final column also widens the position tolerance to 0.5\,m.
It is shown that relaxed reaching tolerances allow PrepTurn to maintain higher speeds through goal transitions, especially for smaller, less demanding turns.
Without a heading requirement, the policy begins turning toward the next goal while still satisfying the current goal's position requirement, which is more clearly visible in the supplementary video.
Further, with a wider position tolerance, the policy also learns to cut corners within the accepted goal regions, shortening the path through the sequence. Sharper turns still require localized slowing, but the policies exploit the available freedom in both path and speed to prepare for subsequent goals.

\begin{table}[t]
\centering
\begingroup
\caption{Failure-adjusted traversal time $t_{\rm FA}$ (s) on the four courses in Fig.~\ref{fig:tolerances}. Values are mean $\pm$ sample SD across three training seeds; every unsuccessful episode is charged the 12\,s horizon. The final column also removes the heading requirement.}
\label{tab:tolerances}
\setlength{\tabcolsep}{3pt}
\footnotesize
\begin{tabular}{@{}lcccc@{}}
\toprule
Course & \begin{tabular}{@{}c@{}}Single goal\\reaching\end{tabular} & PrepTurn & \begin{tabular}{@{}c@{}}PrepTurn\\($\epsilon_\theta=\infty$)\end{tabular} & \begin{tabular}{@{}c@{}}PrepTurn\\($\epsilon_{xy}=0.5$)\end{tabular} \\
\midrule
60\textdegree{} CW & $5.82\pm0.32$ & $4.34\pm0.13$ & $3.96\pm0.15$ & $3.44\pm0.18$ \\
90\textdegree{} CCW & $6.05\pm0.23$ & $4.58\pm0.10$ & $4.29\pm0.10$ & $3.62\pm0.20$ \\
120\textdegree{} zigzag & $5.92\pm0.09$ & $4.71\pm0.14$ & $4.82\pm0.61$ & $4.06\pm0.07$ \\
150\textdegree{} zigzag & $5.88\pm0.20$ & $4.78\pm0.20$ & $4.64\pm0.19$ & $4.12\pm0.08$ \\
\bottomrule
\end{tabular}
\endgroup
\end{table}

These emergent behaviors successfully make traversal times shorter for the full sequence, as shown by the quantitative results in Table~\ref{tab:tolerances} for the same four courses and policy types as Fig.~\ref{fig:tolerances}. Each policy is evaluated over 4096 episodes per training seed on the common rough terrain, with the reaching criteria described above. Removing the heading requirement reduces mean failure-adjusted traversal time on three of the four courses. With the position tolerance additionally widened to 0.5\,m, PrepTurn achieves 14--21\% lower mean failure-adjusted traversal time than standard PrepTurn on every course. These results demonstrate that PrepTurn can learn more efficient sequential traversal under relaxed reaching requirements.

\subsection{Ablation Studies}
\label{sec:reward_factorial}

We examine how reward formulation and next-goal information affect sequential traversal using the two variants in Table~\ref{tab:randomized}. \emph{PrepTurn (simple progress)} replaces the persistent-credit reward with the simple-progress reward while retaining lookahead. \emph{PrepTurn (w/o lookahead)} is trained with the persistent-credit reward, observing only the current goal during both training and evaluation. Each variant is trained from scratch, with all other training and evaluation settings held fixed. Both variants are evaluated under the same protocol as PrepTurn in Section~\ref{sec:setup}.

\noindent\textbf{Reward ablation.}
Simple progress produces faster goal crossings, but this does not translate into better overall task performance. On five-goal sequences, it crosses goals at $2.10$\,m/s compared with PrepTurn's $1.33$\,m/s, yet its fall rate rises from $0.49\%$ to $3.47\%$. These results expose the distinction between passing an individual goal quickly and reliably completing a sequence: higher transition speed alone is not a sufficient measure of useful progress.

This behavior is consistent with the reward design in Section~\ref{reward}. The simple-progress reward encourages immediate error reduction, which may conflict with preparing the next maneuver, while the persistent-credit reward preserves the value of completed goals during subsequent maneuvering and allows more adjustment around goal transitions, but does not amount to the near-stop strategy of single goal reaching.

\noindent\textbf{Lookahead ablation.}
Removing lookahead increases failure-adjusted time from 3.196 to 4.709\,s at $N=2$ and from 6.648 to 10.072\,s at $N=5$. Five-goal success remains high at $98.49\%$, while goal-switch speed falls from 1.330 to 0.721\,m/s. These results support the contribution of next-goal information to efficient sequential local navigation of PrepTurn.

\noindent\textbf{Curriculum ablation.}
When we disable the goal curriculum (Section~\ref{sec:goal_curriculum}) and train from the hardest goal range ($\Delta\theta\in[-\pi,\pi]$, $\ell\in[0,4.5]$) from the start, performance collapses across all goal sequences.
Specifically, both the baseline and PrepTurn frequently fall or get stuck before getting near the goal, achieving near-zero success rates on every sequence. This phenomenon indicates that progressively expanding goal difficulty is critical for learning stable turning behaviors; without curriculum, exploration is insufficient and training becomes stuck at failure.

\subsection{Real-World Deployment}
\label{sec:real}

We deploy the simulation-trained policies on a Unitree Go2 with onboard inference on a Jetson Orin Nano. An invariant extended Kalman filter estimates robot pose from IMU, foot-contact, and joint-position measurements~\cite{barrau2017invariant}. Table~\ref{tab:real_exp} compares traversal times for single goal reaching, PrepTurn, and its relaxed-tolerance variants, reported as mean $\pm$ standard deviation (SD) across 10 trials per policy on each of four turning sequences. All four policies completed every sequence in all trials (100\% success), with no falls observed. PrepTurn, trained with the persistent-credit reward, has a shorter mean traversal time than single goal reaching on all four tasks. Fig.~\ref{fig:teaser} and the supplementary video illustrate the robot motions.

\begin{table}[t]

\centering
\caption{Real-world traversal time (s), reported as mean $\pm$ SD across 10 trials per policy per task.}
\label{tab:real_exp}
\setlength{\tabcolsep}{5.8pt}
\footnotesize
\renewcommand{\arraystretch}{1.08}
\begin{tabular}{@{}lcccc@{}}
\toprule
Goal sequence & \begin{tabular}[c]{@{}c@{}}Single goal\\reaching\end{tabular} & \method{} & \begin{tabular}[c]{@{}c@{}}\method{}\\$(\epsilon_{\theta} = \infty)$\end{tabular} & \begin{tabular}[c]{@{}c@{}}\method{}\\$(\epsilon_{xy} = 0.5)$\end{tabular} \\
\midrule
$60^\circ$ (CW) & $7.12\pm0.24$ & $5.85\pm0.34$ & $5.58\pm0.21$ & $6.01\pm0.19$ \\
$90^\circ$ (CCW) & $7.14\pm0.38$ & $5.98\pm0.23$ & $5.68\pm0.14$ & $5.30\pm0.23$ \\
$120^\circ$ (Zigzag) & $7.19\pm0.37$ & $6.16\pm0.39$ & $6.11\pm0.53$ & $5.48\pm0.55$ \\
$150^\circ$ (Zigzag) & $7.72\pm0.54$ & $6.82\pm0.47$ & $7.21\pm0.39$ & $6.36\pm0.37$ \\
\bottomrule
\vspace{-2.1em}

\end{tabular}
\end{table}

\section{Discussion and Conclusion}

When deploying agile legged robots in the real world, straight-line speed is rarely the true bottleneck; the bottleneck lies in maneuvering. If a robot must decelerate and reorient itself at every waypoint, the advantage gained from sprinting in a straight line is entirely lost. In this work, we introduced PrepTurn and evaluated its ability to navigate sequences of varying goals. By providing the policy with a sequential goal-reaching reward, current-and-next pose goals, and an adaptive curriculum, the robot learns to orchestrate its momentum, proactively adjust its gait, begin turning before reaching a waypoint, and cut corners when tolerances allow. These behaviors were never explicitly rewarded; they arise because the sequential reward structure and lookahead observation create an optimization landscape where anticipatory motion is genuinely advantageous.

A current limitation is that the policy does not automatically adapt to different goal-reaching tolerances: each tolerance setting requires retraining. The behaviors in Fig.~\ref{fig:tolerances} therefore reflect separately learned strategies. In future work, we will investigate conditioning the policy on the tolerances so that a single policy can adjust its approach and turning strategy to different task requirements.

Looking ahead, integrating PrepTurn with global planners and exteroceptive sensing (e.g., LiDAR or vision) would enable the policy to anticipate not only upcoming goals but also obstacles and terrain changes, bringing fast navigation closer to real-world deployment in cluttered and dynamic environments.

\section*{Acknowledgment}
During manuscript drafting and revision, OpenAI Codex was used to help express the authors' intended meaning in academic language throughout the manuscript. All resulting text was rigorously proofread, reviewed, and revised by the authors. Codex also assisted in writing and editing code for the guided velocity tracking baseline (Section~\ref{sec:baseline}) and for trajectory visualization. The authors take full responsibility for the final manuscript and accompanying code.

\bibliographystyle{IEEEtran}
\bibliography{refs}

\appendix[Implementation Details]
\begingroup
\label{app:protocol}

\noindent\textbf{Shared policy architecture.}
All policies use actor and critic MLPs with hidden widths $[512,256,128]$ and ELU activations, and output 12 joint-position targets for PD control at 50\,Hz. PPO uses 1280 parallel Isaac Gym environments with domain randomization. Commands augment the 47 proprioceptive/time features (Section~\ref{sec:background}).

\noindent\textbf{PrepTurn and variants.}
PrepTurn adds six features encoding the current and next pose goals and uses the persistent-credit reward in Eq.~\eqref{eq:sequential_reward}. Single goal reaching adds only the current goal's three features, uses Eq.~\eqref{eq:single_goal_reward}, and switches reached targets at evaluation. Both share the goal curriculum. The simple-progress ablation substitutes Eq.~\eqref{eq:progress_reward}; the no-lookahead ablation removes the next goal. Each ablation and tolerance variant is trained separately.

\noindent\textbf{Guided velocity tracking.}
The PPO tracker instead adds three body-frame velocity commands and learns from sampled commands with velocity-tracking rewards. Its geometric generator is defined in Section~\ref{sec:baseline}; the slowdown weight is $b=u^2(3-2u)$, where $u=\operatorname{clip}(1-d/(0.5\,\mathrm{m}),0,1)$. Goal-directed planar velocity is rotated into the body frame. Yaw rate is $\operatorname{clip}(3\operatorname{wrap}(\theta_g-\psi),-2,2)$\,rad/s, where $\theta_g$ and $\psi$ are target and robot headings. Commands become zero after completion.

\endgroup

\end{document}